\documentclass[10pt,twocolumn,letterpaper]{article}

\usepackage{epsfig}
\usepackage{graphicx}
\usepackage{amsmath}
\usepackage{amssymb}
\usepackage{placeins}
\usepackage{wrapfig}
\usepackage[normalem]{ulem}
\usepackage{biblatex}

\title{Deep multi-frame face super-resolution}
\author{Evgeniya Ustinova, Victor Lempitsky}

\begin{document}

\maketitle

\begin{abstract}
Face verification and recognition problems have seen rapid progress in recent years, however recognition from small size images remains a challenging task that is inherently intertwined with the task of face super-resolution. Tackling this problem using multiple frames is an attractive idea, yet requires solving the alignment problem that is also challenging for low-resolution faces. Here we present a holistic system for multi-frame recognition, alignment, and superresolution of faces. Our neural network architecture restores the central frame of each input sequence additionally taking into account a number of adjacent frames and making use of sub-pixel movements. We present our results using the popular dataset for video face recognition (YouTube Faces). We show a notable improvement of identification score compared to several baselines including the one based on single-image super-resolution.
\end{abstract}

\section{Introduction} Face recognition systems have seen a great progress over the last several years with super-human recognition accuracy attainable in many scenarios. However, the accuracy of recognition degrades very significantly when dealing with very low resolution faces. In such conditions, the tasks of recognition and increasing the effective resolution (\textit{super-resolution}) become intertwined and necessitate joint solution. Indeed, developing  super-resolution techniques without regard for recognition often leads to face \textit{hallucination}, i.e.\ a process that creates plausibly looking faces lacking personal specifics. On the other hand, super-resolution has been known to benefit from recognition for a long time \cite{baker2002limits}.

\begin{figure}[t]
%\begin{wrapfigure}{i}{0.5\textwidth}
\includegraphics[width=\columnwidth]{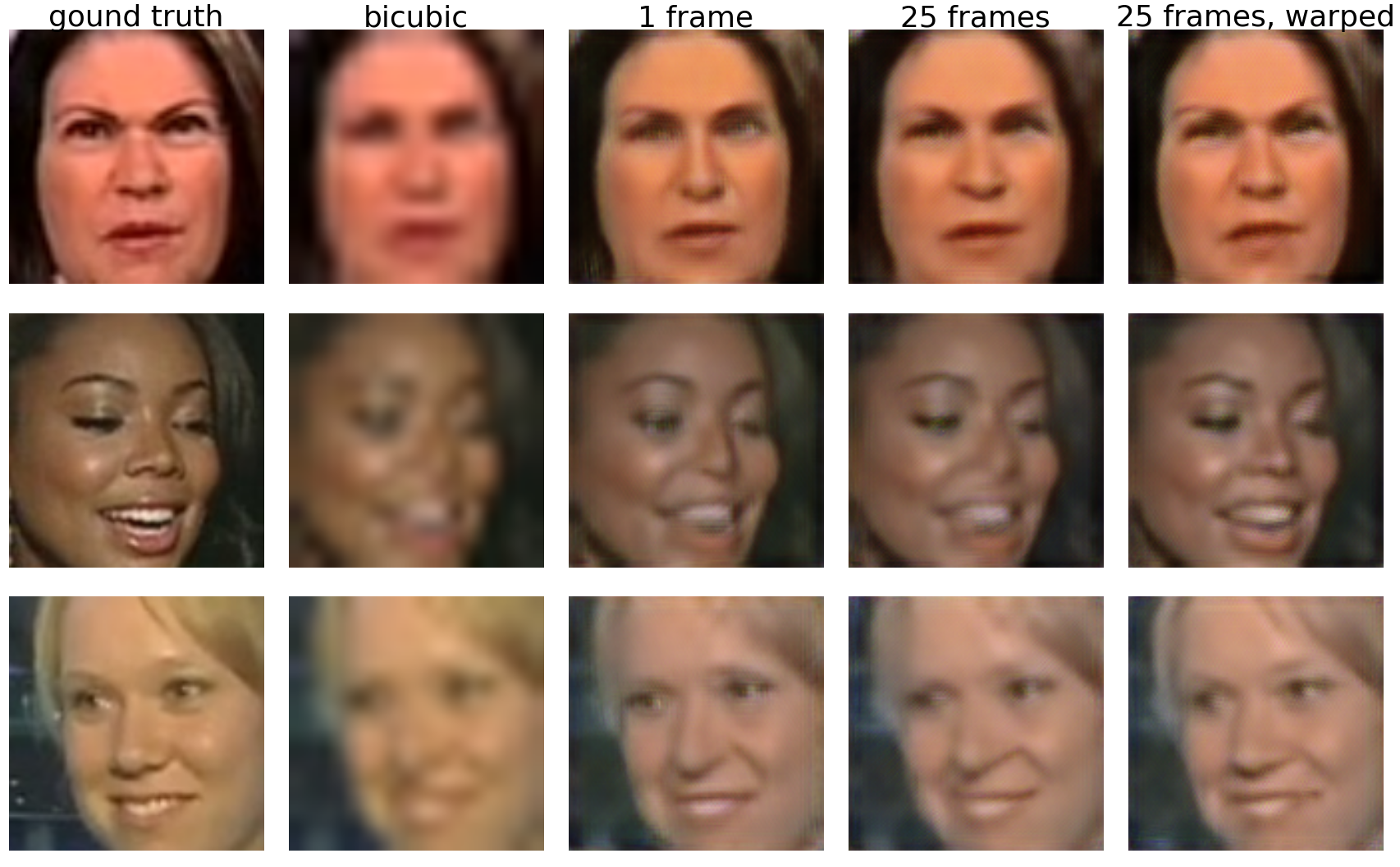}
\caption{Results of different super-resolution convolutional networks on samples from the Youtube Faces (YTF) dataset. From left to right: ground truth, bicubic upsampling,  single-frame superresolution, super-resolution from 25 frames without alignment (ours), super-resolution from 25 frames with warping subnetwork (ours).}
\label{fig:teaser}
%\vspace{10pt}
\end{figure}

While single-image super-resolution has recently drawn considerable attention \cite{ZhuLLT16, TuzelTH16}, super-resolution over large magnification factors can benefit significantly from information accumulated over multiple images, e.g.\ using adjacent frames in a surveillance stream or a video. Traditionally, multi-frame super-resolution has required rigid or non-rigid alignment with sub-pixel accuracy \cite{capel2003computer}. 
At the same time, faces have complex and deformable shapes leading to complex two-dimensional motion patterns which makes motion estimation hard to accomplish at sub-pixel precision. Generally, such precise alignment cannot be accomplished using low-level cues alone, and therefore requires high-level understanding/recognition of face geometry.

Motivated by all these observations, we present a system that performs multi-frame super-resolution by tackling all three inter-related problems, namely super-resolution, non-rigid alignment, and recognition, jointly and simultaneously. The tasks are implemented as modules of a deep neural network architecture that is trained in an end-to-end fashion on a dataset of realistic face videos \cite{WolfHM11}. The forward pass in our network involves pairwise alignment of pairs of frames performed in parallel with feature extraction, while the super-resolution is accomplished by a subsequent reconstruction module that takes warped features of the multiple frames into account. The learning process is driven by a combination of loss functions that includes the recognition-related loss ensuring that the super-resolution process reconstructs person-specific traits.

%Here, we propose and evaluate a deep system that embraces all the aforementioned features: super-resolution, caring for recognition, multiple frames usage, motion recovering and avoiding hallucination. The system is trained end-to-end. \emph{Face Warping} sub-network is responsible for predicting the motion compensation transformation to align pairs of frames in a video-sequence to match one reference frame. The predicted transforms are applied to deep features of the frames in the input sequence, computed with \emph{Frame Feature extractors}. \emph{Combining} sub-network accepts all the transformed frames features and reconstructs the central frame in the sequence. 
Overall, while individual components of our system have been proposed in previous works, to the best of our knowledge, our work is the first that builds a systems that combines face super-resolution, recognition and alignment in a holistic manner. 
We evaluate the proposed architecture on the hold-out part of the YouTube Faces (YTF) dataset (\cite{WolfHM11}). We demonstrate good face verification  performance for the restored images using standard protocols adopted for the YTF dataset. We also show benefits of using multiple frames along with \emph{Face Warping} sub-network over the single-image approach. 
 We additionally compare our approach with  state-of-the-art face hallucination method \cite{ZhuLLT16} and find our method to perform better on the YouTube Faces dataset.

In the remainder of this work, we review the most related approaches in \ref{sec:related} describe the components of the proposed system in \ref{sec:video} and \ref{sec:loss} demonstrate the super-resolution results in \ref{sec:exps} and conclude with a short summary in \ref{sec:summary}. 

\section{Related work}
\label{sec:related}

\subsection{Face super-resolution approaches}

Initially, super-resolution problem has been formulated for low-resolution image sequence that can be used to produced one image of higher resolution. Under this approach, reconstruction constraints are used to make sure that resulting high-resolution image is consistent with the input sequence. 
Maximum a posteriori (MAP) framework has been used in \cite{HardieBA97, SchultzS96, baker2002limits} to take into account such reconstruction constraints along with priors on the high-resolution image.

Precise image registration has been shown to be very important for multi-image \cite{HardieBA97, IraniP91, SchultzS96} and super-resolution methods but is rarely achievable for real  low-resolution data, especially for images depicting non-rigid objects such as faces. In the case of multi-image super-resolution, approximate low-parametric registration could be used instead. For single-image super-resolution, such registration with canonical pose can also be used in order to use stronger face-specific priors \cite{LiuSF07}.

Having assumed ideal image registration, Baker \emph{et al.} \cite{baker2002limits} demonstrate the effect of introducing additional recognition-based prior. The authors include it into the task formulation as additional constraints based on particular examples from  training set chosen by similarity to the regions of the input image.  Later, more advanced techniques for modeling such face-specific constraints were introduced in \cite{LiuSF07} for single-image face super-resolution. The authors decomposed such constraints to \emph{global} and \emph{local} constraints. A similar idea was also adopted in \cite{TuzelTH16} to build a ConvNet architecture for single-image face super-resolution.

%TODO : maybe elaborate on this.
%correspondence between low res and high-res patches 
%dictionary learning: faces \cite{YangWLCH12, TianT16}, direct patch correspondence :  \cite{MaZQ10}, methods are also very dependent on alignment, general and CNN :  \cite{DongLHT16} 

\subsection{Deep learning for face super-resolution}

Several recent deep learning approached \cite{TuzelTH16,ZhuLLT16} focus on single-image face super-resolution problem.
Tuzel \emph{et al.} \cite{TuzelTH16} suggested the end-to-end learning scheme to obtain high quality results even in the case of 8x downsampling. The architecture includes two parts corresponding to global and local modeling of the face features. In the \emph{global} network, the fully-connected layers are used to capture the global face structure, while the \emph{local} part consists of convolution layers that are used to model local face features. Additionally, authors apply adversarial learning to achieve more realistic results. 
Below, we use the \textit{Perceptual loss} for the similar goal.

Zhu \emph{et al.}~\cite{ZhuLLT16}  proposed a cascaded scheme that includes iterative high-resolution image refinement and pose estimation from these refined images. This approach is motivated by the fact that face super-resolution and face pose estimation tasks are related, and it is easier to increase image resolution knowing the pose of the face and vice versa. Special gated architecture is introduced to effectively combine high-frequency and low-frequency information and make use of face pose estimated at previous scale levels.
All the aforementioned methods only consider single-frame face super-resolution/restoration. To the best of our knowledge, none of the recent works on deep face super-resolution consider multi-frame scenario.

\subsection{Restoration and recognition}
%\cite{Hennings-YeomansBK08, ZhangYZNH11}
%The idea of combining the two tasks of face image restoration and face recognition has been investigated in several works

Initially, recognition was incorporated into face super-resolution in the form of face-specific priors \cite{baker2002limits, LiuSF07}. Several works considered an explicit combination of face recognition and face restoration tasks: \cite{Hennings-YeomansBK08, ZhangYZNH11}.
The proposed approaches perform recognition using a limited number of gallery images. In parallel, they reconstruct the input image and classify it based on labels of the most relevant examples found in the train set. Here we work in a different setting using a feed-forward deep neural network to produce the restored high-resolution image without using the gallery set at test time.

\subsection{Deep learning for video-based super-resolution}
As mentioned earlier, video data bring more useful information compared to isolated frames and can be used to enhance the restoration algorithms. Naturally, most recent deep learning approaches, which focus on video super-resolution \cite{LiaoTLMJ15, SongDQ16, TaoGLWJ17}, use motion estimation to align a number of subsequent frames to make use of sub-pixel motion and reveal more object details. Liao~\emph{et al.}~\cite{LiaoTLMJ15} use different optical flow methods to generate super-resolution "drafts". Such drafts can then be further combined into the final reconstruction using a number of convolutional layers. Kappeler~\emph{et al.}~\cite{KappelerYDK16} also experiment with different variants of video-based super-resolution architecture incorporating neighbouring frames alignment. 

%here it is interesting to check if Warping Subnet even changes during training => it would be possible to say that we benefit from end-to-end learring in this case.
Our idea is to adopt similar approach based on video-data and frame alignment for human faces. Importantly, we aim not only at enhancing image quality but also at preserving face identities and therefore improving face verification quality for the restored images.

%\begin{figure*}
%\begin{center}

%\includegraphics[width=\textwidth]{images/method/vgg_loss.pdf}

%\caption{General scheme of learning super-resolution with perceptual loss. Along with L2 loss, the \textit{Perceptual loss} \cite{JohnsonAF16} is used, that is the loss computed for high-level features extracted from pre-trained CNN. The weights of the pre-trained CNN are fixed during training. See \ref{sec:vgg_loss} for the details.}

%\label{fig:vgg_loss}

%\end{center}

%\end{figure*}

\begin{figure*}
%\begin{center}

\includegraphics[width=\textwidth]{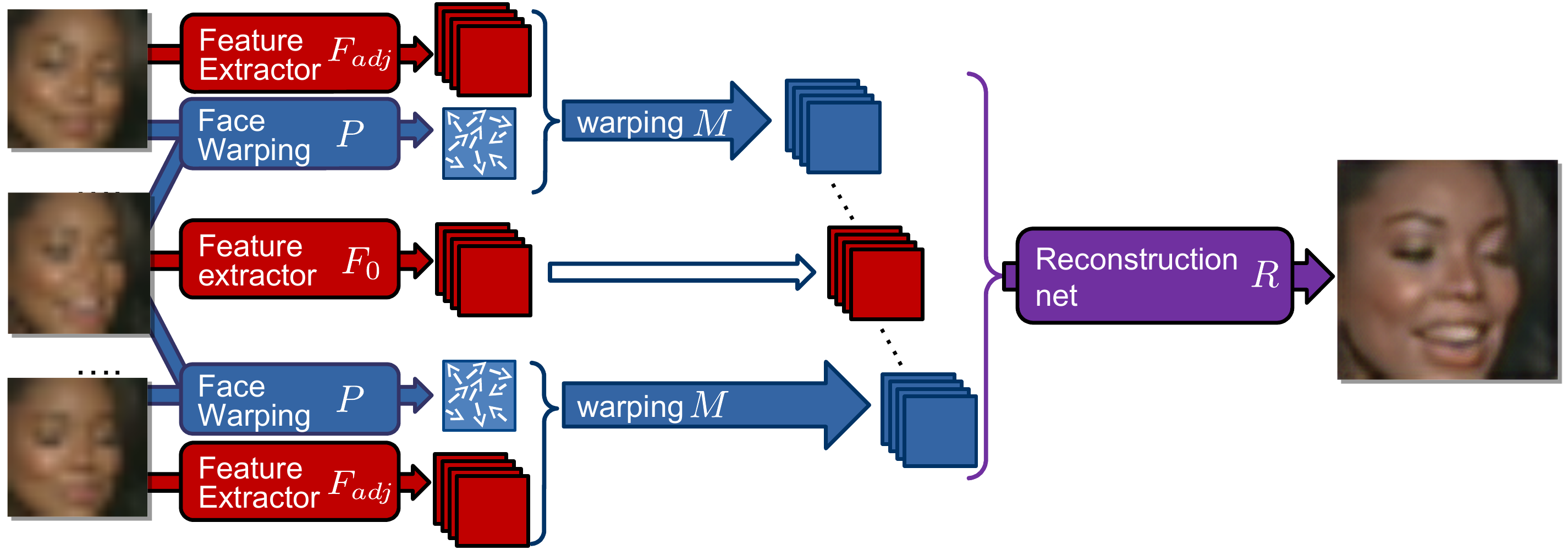}

\caption{The general scheme of our video-based super-resolution convolutional network. \emph{Feature Extractor} modules (red) compute features for every frame in the input sequence, which are then warped by the \emph{Face Warping} sub-network (blue). The transformed feature maps are fed into the \emph{Reconstruction} sub-network (purple) that produces the central frame reconstruction. Spatial transformer modules \cite{JaderbergSZK15} with thin plate spline transformer are used in the \emph{Face Warping} sub-network.}

\label{fig:video}

%\end{center}

\end{figure*}

\section{Multi-frame face super-resolution}
%scheme
\label{sec:video}
Here we describe the proposed multi-frame face super-resolution convolutional neural network. Our architecture consists of three main learnable modules:
\begin{enumerate}
\item\emph{Feature Extractor} sub-network that computes features for every frame in the input sequence, \item\emph{Face Warping} sub-network that aligns input frames with the central frame,  
\item\emph{Reconstruction} sub-network that accepts warped features of input frames and perform reconstruction of the central frame in the sequence.
\end{enumerate}
The modules are combined as follows (\ref{fig:video}). \emph{Feature Extractor} outputs features for every frame in the input sequence. The feature maps are then warped by the \emph{Face Warping} sub-network. The transformed feature maps are fed into the \emph{Reconstruction} sub-network that produces the central frame reconstruction. Another option was to warp all the faces to match one standard pose, but as central frame warping may introduce additional distortions and loss of visual information, we only perform pairwise alignment between the central frame and other frames in the input sequence. 
%TODO can we actually perform warping??

In more detail, our goal is to restore the central frame $s_{0}$ of the input low-resolution sequence $s = [s_{-k}, ..., s_{0}, ..., s_{k}]$ of length $2k+1$. Here we work with RGB-images: $s_{i}\subseteq \mathbb{R}^{3\times h \times w}$, $i\subseteq [-k,..,k]$. We then start by processing each input frame $s_{i}$ with the feature extractor sub-networks $F_{i}$:
\begin{equation}
 f_{i} = F_{i}(\theta_{s_{i}}; F_{i}), 
 f_{i}\subseteq \mathbb{R}^{D\times {h}' \times {w}'}
\end{equation}
where $D$ is the number of output feature maps of size ${h}'\times{w}'$. Here all the feature extractors $F_{i}$ share parameters $\theta_{F_{i}}$, except for the feature extractor for the central frame $F_{0}$: $\theta_{F_{i}} == \theta_{F_{adj}}$, $F_{i} == F_{adj}$, where $i\subseteq [-k,..., -1, 1, ..., k]$. This is done for simplicity, and to avoid excessive memory consumption for long video-sequences.

To align all the adjacent frames $s_{i}$,  $i\subseteq [-k,..., -1, 1, ..., k]$, with the central frame $ s_{0}$, we incorporate the \emph{Face Warping} sub-network that consists of the learnable warping predictor $P(s_{0}, s_{i};\theta_{P})$ and the warping module $M(p_{i}, f_{i})$ that accepts transform parameters computed by $P$ and feature maps $F_{i}$. $M$ performs warping using differentiable sampling introduced in \cite{JaderbergSZK15}.

The warping predictor $P$ accepts frame pairs and outputs transform parameters $p_{i}$  that are used to warp each frame $s_{i}$, $i\neq0$, in the input sequence: 
\begin{equation}
    p_{i} = P(s_{0}, s_{i};\theta_{P}),
\end{equation}
where $i\subseteq [-k,..., -1, 1, ..., k]$.

Feature maps $f_{i}$ are transformed using warping parameters $p_{i}$ in differentiable manner \cite{JaderbergSZK15}: 
\begin{equation}
    f_{i}^{M}= M(p_{i}, f_{i}) = S(\mathcal{T}_{p_{i}}(G)), 
\end{equation}
where $i\subseteq[-k,..., -1, 1, ..., k]$, $\mathcal{T}$ is a predefined transform applied to the regular grid $G$ \cite{JaderbergSZK15}. Here we use Thin Plate Spline transform that is suitable for modeling non-rigid deformations inherent to face images. Therefore, $p_{i}\subseteq  \mathbb{R}^{2C\times 1 }$, where $C$ is a number of anchor points used for warping.

All the warped feature maps $f_{i}^M$ along with central frame feature map $f_{0}$ are then stacked together:
\begin{equation}
    f_{stacked}^{M} = [f_{-k}^{M}, ...,f_{0}, ..., f_{k}^{M}],
\end{equation}
 $f_{stacked}^{M} \subseteq  \mathbb{R}^{(2k+1)D\times {h}' \times {w}'}$.
 The resulting features $f_{stacked}^{M}$ are fed into \emph{Reconstruction} sub-network $R$, resulting in restored image $s_{N}^{R}$:
 \begin{equation}
 s_{0}^{R}=R(f_{stacked}^{M}; \theta_{R})
\end{equation}

The described architecture is learned in a supervised manner, using the loss function calculated for the ground truth image $s_{0}^{G}$ and the restored image $s_{0}^{R}$. The loss function is discussed in details in the section \ref{sec:loss}.

\section{Perceptual Loss for face super-resolution}

\label{sec:loss}
Perceptual losses have already been used for general super-resolution \cite{JohnsonAF16} and face frontalization \cite{cole2017face}. The idea is to compare high-level features for the ground truth image and the restored image in addition to pixel-level data. Here we use pre-trained VGG-face \cite{ParkhiVZ15} model (weights are fixed during training) to extract such features. The motivation behind such approach is that, first, less blurry results can be achieved \cite{JohnsonAF16}. Second, we use the pre-trained verification CNN for feature extraction, so there is a hope that our super-resolution neural network will be able to focus on facial features that are important for face identification. 
%The general scheme of the corresponding CNN 
%architecture (similar to \cite{JohnsonAF16}) is shown in Figure \ref{fig:vgg_loss}. 

%In general case there may be several perceptual loss terms that are added to form the final loss:
Our neural network described in the section \ref{sec:video} in learned using the following objective that includes pixel-level term along with sum of a number of feature-level terms:
\begin{equation}
\label{eq:loss}
\begin{aligned}
    L_{\theta_{F_{0}}, \theta_{F_{adj}}, \theta_{P}, \theta_{R}}(s_{-k}, ..., s_{0}, ..., s_{k}, s^{G}) = \\ \left \| s_{0}^{G} - s_{0}^{R}\right \|^{2}_{2} + \sum_{l \subseteq {layers}}  {\lambda_l \left \| s_{0,l}^{G} - s_{0,l}^{R}\right \|^{2}_{2}}
\end{aligned}
\end{equation}
where $s_{0}^{G}$ and $s_{0}^{R}$ are ground truth image and reconstruction for the frame $s_{0}$, $s_{0, l}^{G}$  and $s_{0, l}^{R}$ the features extracted by the layer $l$ of the ground truth image  $s_{0}^{G}$ and the restored image  $s_{0}^{R}$ accordingly, $\lambda_l$ is a fixed weight assigned to the corresponding loss components. Unlike \cite{cole2017face}, we use mid-level features extracted using VGG-face model as we observed that this leads to better results than using last fully connected layer of VGG-face.
\section{Experiments}

\label{sec:exps}
\subsection{Datasets}

In this work two popular face verification datasets were used. We use Labeled Faces in the Wild (LFW) to evaluate the \textit{baseline architecture}, which closely follows \cite{TuzelTH16} but is trained using \emph{Perceptual loss} (see section \ref{sec:loss} for details). Our multi-frame architecture described in section~\ref{sec:video} is evaluated using YouTube Faces dataset. The ablation study on the same dataset include evaluation of different architecture variants described in section~\ref{sec:arch}.

Labeled Faces in the Wild (LFW) \cite{KaramZ15} contains 13,233 images with 5,749 identities. The standard evaluation procedure for LFW includes similarity estimation for the given set of image pairs. All pairs are split into ten subsets non-overlapping in terms of identities. We use standard test subset $1$ to evaluate our super-resolution methods. Identities that are not included into test subset $1$ are thus used for training. To evaluate the recognition quality, we use the Equal Error Rate (EER) (i.e.\ the error rate at the ROC operating point where the false positive and false negative rates are equal). To get EER values, we first  compute descriptors for the face images using pre-trained face recognition model \cite{ParkhiVZ15}.
%TODO : how many persons in train/test

%EER does not depend on the similarity threshold that separates similarity values for 'positive' and 'negative' pairs of images. 

To evaluate our video-based super-resolution method, we use the YouTube Faces (YTF) dataset \cite{WolfHM11}. It contains 3,425 videos of 1,595
people collected from YouTube, with an average of 2 videos per identity. Similarly to LFW, EER is used to evaluate recognition quality for the YTF dataset. The only difference is that the similarity value is estimated for each pair of videos. To do this, we calculate the mean similarity for all pairs of frames of the two tracks. As this is time-consuming operation, we take only first $100$ frames in each video (the first $20$ frames were used for comparison with \cite{ZhuLLT16}). %The number of the frames used for the evaluation is less because we only take those frames that at least have $12$ neighboring subsequent frames before and after them. This is done to compare results for different variants of the video-based super-resolution method.

We also use PSNR (for luminance channel) and Euclidean distance metrics for comparing different variants of model described in \ref{sec:video}. We use $623$ images from the YTF dataset to compare PSNR and Euclidean distance between features of ground truth and reconstructed images extracted using VGG face model layers \emph{pool3}, \emph{pool4} and \emph{fc7}.

All experiments were performed by considering the downsampled versions of images from the described datasets as inputs. Unless specified otherwise, we give results are for $\times$8 magnification factor as it is a very challenging setting, which still allows to see meaningful variations between methods (much higher magnification factors result in all methods performing equally bad, and much lower magnification factors result in all methods performing equally well).

\subsection{Architectures}

\label{sec:arch}
The following architectures were used in our experiments:
\begin{enumerate}
    \item The single-image architecture from \cite{TuzelTH16} denoted \texttt{f$1$} ( \emph{8$\times$GN} and  \emph{8 layer LN} were used for \emph{Local} and \emph{Global} modules of the architecture).
    \item The multi-frame architecture similar to described in Section~\ref{sec:video} but without \emph{Face Warping} subnetwork. This architecture stacks features for all frames  without alignment and then processes the stack by the \emph{Reconstruction} subnetwork. We experiment with two variants of this architecture \texttt{f$5$} and \texttt{f$25$} that accept sequences of length $5$ and $25$ accordingly.
    \item The full multi-frame architecture described in Section~\ref{sec:video} is also implemented in two variants \texttt{f$5$warp} and \texttt{f$25$warp} that accept sequences of length $5$ and $25$ accordingly.
\end{enumerate} 
The single-image architecture \cite{TuzelTH16} is evaluated for the LFW dataset and several modes of using \emph{Perceptual} loss are compared in the section \ref{sec:perceptual}.
The proposed multi-frame architecture (section \ref{sec:video}) is evaluated on the YTF dataset.
Further details of the architectures are discussed below.

\subsection{Training the model}

The \emph{feature extractor} sub-network is implemented similar to \emph{8$\times$GN} in \emph{Global} module in \cite{TuzelTH16}. It includes two parallel streams. The first stream performs upsampling using one deconvolution layer, while the second stream consists of four fully connected layers. The first three fully connected layers have $256$ neurons, while the last one has the size $128\times128$.

The \emph{face warping} sub-network accepts pairs of images. Initially, the two images are processed separately using  two sub-networks consisting of three convolution layers with $20$ filters of sizes $3\times3$, $3\times3$ and $1\times1$ accordingly. The outputs of these two streams are concatenated and passed to the next five convolution layers with $100$ filters of size $3\times3$ each.
Then three fully connected layers with $256$ neurons are applied. The final fully connected layer then outputs shifts for coordinate pairs of $64$ control points. The warping is performed using thin plate spline with $64$ control points located in the nodes of the regular grid \cite{JaderbergSZK15}.

The \emph{reconstruction} sub-network is implemented similarly to \emph{Local} module in \cite{TuzelTH16}. It consists of eight convolution layers:
\vspace{5mm}
\begin{tabular}{c |c c c c c c c c }
\hline
  \#conv & 1& 2& 3 & 4 & 5&6  & 7 &8 \\
  \#filters & $16$ & $32$&$64$&$64$&$64$&$32$&$16$&$3$  \\
  filter size &$5$& $7$& $7$& $7$& $7$& $7$& $5$& $5$ \\
\hline
\end{tabular}

ReLU (rectified linear unit) activation functions are used for all the layers except the last convolution layer in the \emph{Reconstruction} sub-network. The latter uses the sigmoid activation.

The architecture is trained using \ref{eq:loss} and the ADAM optimizer~\cite{KingmaB14} for ~1,500 epochs with learning rate fixed to $1e-4$. Batch size is set to $10$. Pre-trained VGG-16 face model from \cite{ParkhiVZ15} is used for calculating deep features within loss \ref{eq:loss}.

The \emph{face warping} sub-network is pre-trained in unsupervised manner using pairs of images. The L2 loss for the warped and reference images is used for training.

For the LFW dataset, we used $\lambda_{pool3}$, $\lambda_{pool4}$ and $\lambda_{fc7}$ set to $10^3$ for the loss \ref{eq:loss}. For the YTF dataset, the value of $10^5$ was used because the quality of ground-truth images in YTF is much worse than for LFW and therefore pixel-level loss is less useful for the training on YTF. Increasing values of $\lambda_{pool3}$ and $\lambda_{pool4}$ also affects the color of resulting images, as deep features are more color-independent (c.f.~\ref{fig:ytube}, row $7$, where the green background became grey in the reconstruction).

\subsection{Perceptual Loss effects}
\label{sec:perceptual}

First, we experimented with training the baseline single-image super-resolution ConvNet (the architecture from \cite{TuzelTH16}) using the Perceptual loss in addition to the pixel-level loss. Here we show some of the results that demonstrate the effect of using deep features extracted from different layers of the pre-trained VGG-face model and choose the most beneficial variant.

As mentioned above, we evaluate the models using the split 1 of the LFW dataset. All images were resized to $128\times128$ pixels, blurred (Gaussian kenel, $\sigma=2.4$) and downsampled to $16\times16$ pixels for training and testing.

%It turned out that using fully-connected layers of the pre-trained VGG-face model leads to the decreasing image quality. See Figure \ref{fig:fc7} for results of learning with perceptual loss using \emph{fc7} layer of the VGG-face model additionally to the pixel-level L2 loss. 

\begin{figure}[t]

\includegraphics[width=\columnwidth]{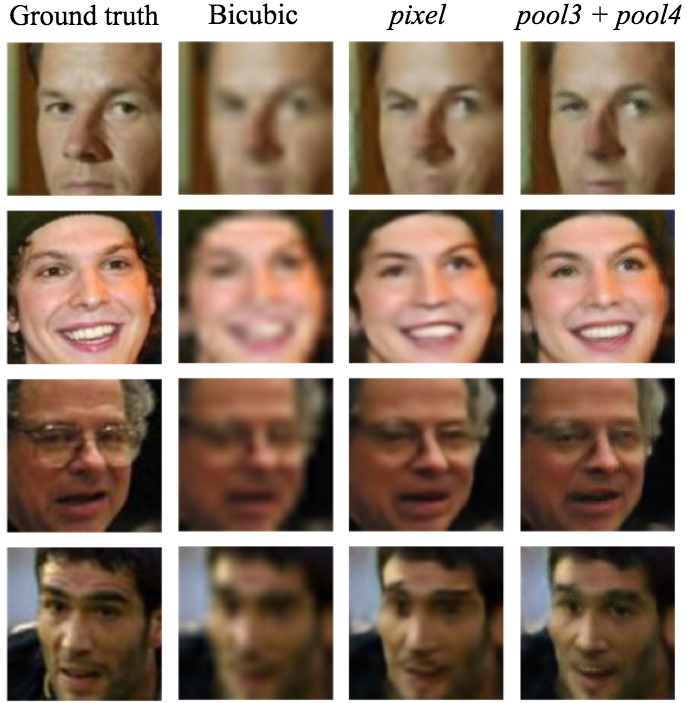}

\caption{The results achieved using additional \emph{Perceptual} loss terms for layers \emph{pool3} and \emph{pool4} compared with result for pixel-level loss and bicubic interpolation. We can see the clear improvement when using layers \emph{pool3} and \emph{pool4} along with pixel-level data.}

\label{fig:layers}

\vspace{10pt}
\end{figure}

The most visually plausible results were achieved by using the \emph{pool3} and \emph{pool4} layers for learning along pith pixel-level data. The results on LFW hold-out set were compared with bicubic interpolation as well as with the variant learned with pixel-level loss only. The comparison in Figure \ref{fig:layers} shows clear improvement of the system that uses pixel-level loss over the baselines. 

We also observed that using perceptual loss improved the recognition score. Recognition scores ($100\%$-EER) for these cases are compared in  \ref{tab:lfw_score}. At the same time, as expected, best results in terms of PSNR (peak signal-to-noise ratio) were achieved with pixel-level loss. %TODO : insert values

%Additionally, we have compared the results for down-sampling factor $4$ (input images were of size $32\times32$), but there was no difference in either visual quality or recognition score between the results \emph{pixel} and \emph{pixel} $+$ \emph{pool3} $+$ \emph{pool4} modes.

\begin{table*}
\centering
\begin{tabular}{ c | c c c c c }
  & gt & bicubic & \emph{pixel}  & \emph{pixel} + \emph{pool3} + \emph{pool4} & \emph{pixel} + \emph{fc7}  \\
\hline
train & -      & -    & 85.56 & 86.19 & 90.74\\
test  & 95.33 & 73.01 & 83.02 & \bf{85.02} & 84.33\\

\end{tabular}
\caption{100\% - EER (Equal error rate) values for the standard split $1$ of the FLW dataset. The scores for bicubic upsamling and ground truth images are also included. \emph{pixel} - the results achieved using pixel-level loss only. Best results were achieved using pixel-level loss along with \emph{Perceptual} loss for layers \emph{pool3} and \emph{pool4} of the pre-trained VGG-face model.}
\label{tab:lfw_score}
\end{table*}

% \begin{figure*}
% \begin{center}

% \includegraphics[width=\textwidth]{images/lfw/fc7.png}

% \caption{The results achieved by training the super-resolution network using additional perceptual loss term that compares reconstruction and ground truth images representations extracted from the \emph{fc7} layer of the pre-trained VGG-face model. No image quality improvement is observed, on the contrary, images become less realistic with the increase of perceptual loss weight. }

% \label{fig:fc7}

% \end{center}

% \end{figure*}

\subsection{Experiments with video-data}

We have performed experiments on video-based face super-resolution using different variants of face super-resolution network described in \ref{sec:video}. All images were fist resized to $128\times128$ pixels, blurred with the Gaussian kernel ($\sigma = 2.4 $), and downsampled to the size $16\times16$. 
Here we use previously chosen perceptual loss training mode \emph{pixel} + \emph{pool3} + \emph{pool4} that uses pixel-level differences along with differences of features at \emph{pool3} and \emph{pool4} layers to compute the final loss value. The weight equal to $10^5$ was used for both Perceptual loss components  \emph{pool3} and \emph{pool4}.

In \ref{tab:ytube_score} $100\%-$EER(Equal Error Rate) value for single-image super-resolution is compared to video-based results using different number of adjacent frames with and without frame alignment (warping). Values for ground truth images (without distortions) and bicubic upsampling are also included. Note, that ground truth performance is less than reported in \cite{ParkhiVZ15} for several reasons: we use only one scale $256\times256$ and only one crop (central) to calculate face image descriptor. Moreover, all ground truth images were resized to $128\times128$ pixels to make fair comparison with our super-resolution method that outputs images of this size. 

\begin{figure*}
\begin{center}

\includegraphics[width=0.75\textwidth]{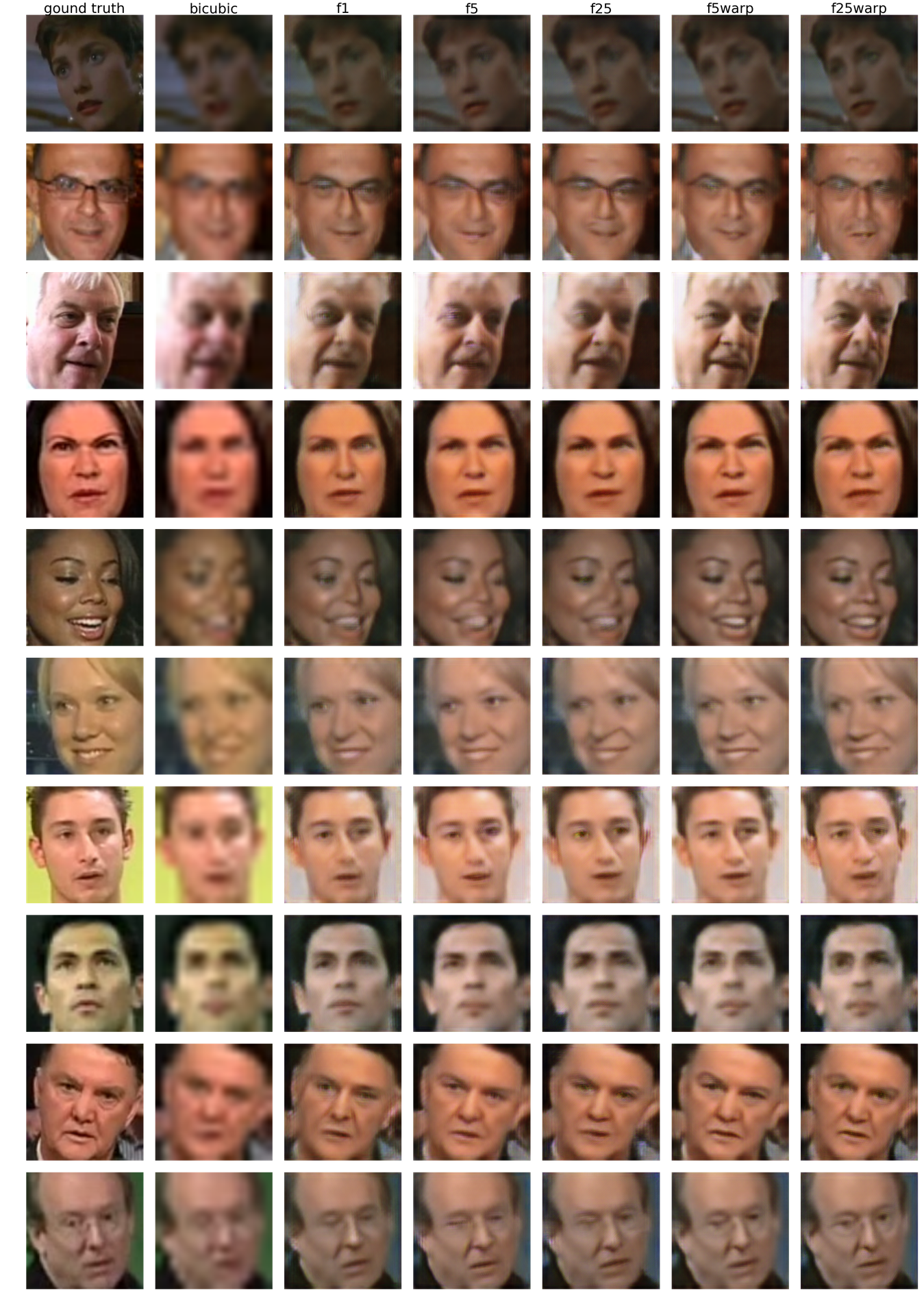}

\caption{Results of different super-resolution CNNs for some of the images in the YTF dataset. Ground truth images and bicubic upsampling results are in the first two columns. See sections \ref{sec:arch} and \ref{sec:video} for details.}

\label{fig:ytube}

\end{center}

\end{figure*}

The best-performing method was \texttt{f$25$warp} that uses 25 frames (one central and $24$ neighboring ones) and frame warping. Using several frames is generally better than using only one frame (c.f.~super-resolution results comparison in \ref{fig:ytube}). In \ref{tab:ytube_score} we can also see that the architecture \texttt{f$25$warp} gives reconstruction that are the closest to ground-truth when comparing images using L2 distance between their VGG deep features. As expected, the architecture \texttt{f$25$warp} gives the best PSNR value. In general, we can see uniform improvement in all used metrics for architectures that use longer sequences in comparison to architectures that use less frames. Analogously, there is improvement for architectures that include \emph{Face Warping} module (\texttt{f$25$warp} and \texttt{f$5$warp}) in comparison to architectures without \emph{Face Warping} module (\texttt{f$25$} and \texttt{f$5$}).

\subsection{User study for multi-frame face super-resolution}
\label{sec:ustudy}

We also carried out a used study in order to compare human perception of images produced by architectures \texttt{f$1$}, \texttt{f$25$} and \texttt{f$25$warp}. 
We thus compared the following three pairs of architectures: \texttt{f$25$} vs \texttt{f$25$warp}, \texttt{f$1$} vs \texttt{f$25$warp}, \texttt{f$1$} vs \texttt{f$25$}. The comparison interface presented five result pairs (for each of the three pairs of methods) from the YTF dataset to each user. Within the pair, the order was randomized. Between the pair, we showed a full-resolution frame from the same sequence (different from the central frame). The users were asked to pick one of the two reconstruction that (a) resemble the reference image more and (b) has better overall quality. The users were asked to prioritize resemblance over visual quality. Finally, if the two images seemed to the user absolutely on par both in terms of resemblance and quality, the user were allowed to click on the middle (reference) frame to indicate this. We however asked the users to abstain from this as much as possible. Overall $42$ users took part in the study.

%Second criterion (less important) was the overall image quality. In addition to two images that can be chosen, users could also answer "none", meaning that it is hard to decide which of the images is better.

\begin{figure}[t]

\begin{center}

\includegraphics[width=\columnwidth]{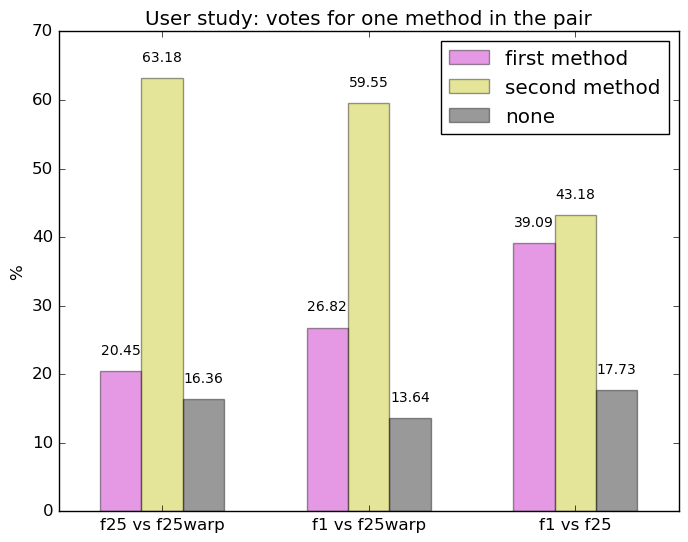}
\end{center}
\caption{Results of user study for the three pairs of architectures evaluated in this works. See details in section \ref{sec:ustudy}}

\label{fig:ustudy}

\vspace{10pt}
\end{figure}

The results of the user study (\ref{fig:ustudy}) showed that it is easy to tell the difference between the results of  \texttt{f$25$} vs \texttt{f$25$warp}  and \texttt{f$1$} vs \texttt{f$25$warp}. It appeared to be challenging to differ \texttt{f$1$} vs \texttt{f$25$} methods. These observations indicate that \emph{Face Warping} module is essential for obtaining better results. Using more frames without warping thus did not bring advantage over single-frame super-resolution.

\subsection{Comparison to state of the art}

Here we compare our method with one of the latest single-image super-resolution methods \cite{ZhuLLT16}. As we use different format of input images than in \cite{ZhuLLT16} (for example, we use tighter crops for faces), we evaluate \cite{ZhuLLT16} on the YTF dataset using less magnification factor than we use in this work. For our methods we use inputs of size $16\time16$, while using $32\times32$ for \cite{ZhuLLT16}. We use the code published for \cite{ZhuLLT16} for the evaluation and therefore ensure that all the inputs are preprocessed as appropriate for \cite{ZhuLLT16} (including face detector for cropping the ground truth images from initial frames). As in \cite{ZhuLLT16}, the downsampled images were created without applying gaussian blur beforehands, we perform the comparison on such images after retraining our architecture for such regime. 

We observed that our method performs better than \cite{ZhuLLT16} in terms of $100$\%-EER metric: 85.14 (ours) vs 82.32 (\cite{ZhuLLT16}). Also, we can see that $100$\%-EER value for \cite{ZhuLLT16} is very close to the $100$\%-EER value for \texttt{f$1$} architecture that we use as a single-image baseline in this work (see Table \ref{tab:ytube_score}).
%TODO maybe insert L2 varues, but we need to change the code in \cite{ZhuLLT16} so that it wouldnt rotate the images.

\begin{table*}
\centering
\begin{tabular}{c | c c c c c c c c}
 metric & gt & bicubic & \texttt{f$1$} & \texttt{f$5$} &  \texttt{f$25$} & \texttt{f$5$warp} &  \texttt{f$25$warp}  \\
\hline
$100$\%-EER & 85.60 & 78.80 & 82.40 & 83.20 & 83.60 & 84.39 & \bf{85.20}\\
PSNR      && & 27.64 & 28.22 & 28.28 &  \bf{29.17} & 29.12 \\
L2, \emph{pool3} &&& 0.6512 & 0.6199 & 0.6295 & 0.5568 &  \bf{0.5542} \\
L2, \emph{pool4} &&& 0.7110 & 0.6814 & 0.6918 & 0.6126 &  \bf{0.6074} \\
L2, \emph{fc7} &&& 0.4828 & 0.4478 & 0.4583 & 0.3811 &  \bf{0.3695} \\

\end{tabular}
\caption{Different metrics for several architectures evaluated in this work (see section \ref{sec:arch}). First row (face verification metric): 100\%-EER(Equal error rate), second row: PSNR - peak signal to noise ratio. Rows $3$-$5$ show mean Euclidean distances between deep features of ground truth and reconstructed images extracted using VGG-face layers \emph{pool3}, \emph{pool4}, \emph{fc7} (after normalization). Such metrics are strongly correlated with recognition. The best performing architecture is \texttt{f$25$warp} that accepts sequences of length $25$ and includes \emph{Face Warping} sub-network. See section \ref{sec:video} for details.}
\label{tab:ytube_score}
\end{table*}

\section{Summary}
\label{sec:summary}
In this work we present a deep neural network for multi-frame face super-resolution that performs alignment, reconstruction and recognition in holistic manner and learns respective modules in the end-to-end fashion. We evaluate different variants of the proposed architecture including single-image baseline. In the experiments on YouTube Faces dataset, we demonstrate the advantages of having the alignment module in the system. In the presence of alignment, we observe the improvement both in visual quality (confirmed by the user study) and in the face recognition accuracy over the single frame baselines and the single frame system \cite{ZhuLLT16}.

\medskip
 
\printbibliography

\end{document}